\documentclass[lettersize,journal]{IEEEtran}

\usepackage{amsmath,amsfonts}
\usepackage{algorithmic}
\usepackage{algorithm}
\usepackage{array}
\usepackage[caption=false,font=normalsize,labelfont=sf,textfont=sf]{subfig}
\usepackage{textcomp}
\usepackage{stfloats}
\usepackage{url}
\usepackage{verbatim}
\usepackage{graphicx}
\usepackage{cite}

\usepackage{multirow}
\usepackage{booktabs}
\usepackage{graphicx}
\usepackage{amsmath}
\usepackage{amssymb}
\usepackage{array}
\usepackage{colortbl} 
\usepackage{xcolor}   
\usepackage{arydshln}
\begin{document}

\title{SynMind: Reducing Semantic Hallucination in fMRI-Based Image Reconstruction}

\author{Lan~Yang,
        Minghan~Yang,
        Ke~Li,
        Honggang~Zhang,~\IEEEmembership{Senior Member,~IEEE,}
        Kaiyue~Pang, and
        Yi-Zhe~Song,~\IEEEmembership{Senior Member,~IEEE}
\IEEEcompsocitemizethanks{\IEEEcompsocthanksitem Lan Yang, Minghan Yang, Ke Li,Honggang Zhang are with the School of Artificial Intelligence, Beijing University of Posts
and Telecommunications, 100876, China. \protect E-mail: \{ylan, yangminghan, like1990, zhhg\}@bupt.edu.cn.
\IEEEcompsocthanksitem Kaiyue Pang and Yi-Zhe Song are with SketchX Lab, Centre for Vision, Speech and Signal Processing (CVSSP), University of Surrey, GU2 7XH, United Kingdom. \protect E-mail: thatkpang@gmail.com, y.song@surrey.ac.uk.}
\thanks{Corresponding Author: Honggang Zhang}}

\markboth{Journal of \LaTeX\ Class Files,~Vol.~14, No.~8, August~2021}%
{Shell \MakeLowercase{\textit{et al.}}: A Sample Article Using IEEEtran.cls for IEEE Journals}

\IEEEpubid{0000--0000/00\$00.00~\copyright~2021 IEEE}

\maketitle

\begin{abstract}
Recent advances in fMRI-based image reconstruction have achieved remarkable photo-realistic fidelity. Yet, a persistent limitation remains: while reconstructed images often appear naturalistic and holistically similar to the target stimuli, they frequently suffer from severe semantic misalignment -- salient objects are often replaced or hallucinated despite high visual quality. In this work, we address this limitation by rethinking the role of explicit semantic interpretation in fMRI decoding. We argue that existing methods rely too heavily on entangled visual embeddings which prioritize low-level appearance cues -- such as texture and global gist -- over explicit semantic identity. To overcome this, we parse fMRI signals into rich, sentence-level semantic descriptions that mirror the hierarchical and compositional nature of human visual understanding. We achieve this by leveraging grounded VLMs to generate synthetic, human-like, multi-granularity textual representations that capture object identities and spatial organization. Built upon this foundation, we propose SynMind, a framework that integrates these explicit semantic encodings with visual priors to condition a pretrained diffusion model. Extensive experiments demonstrate that SynMind outperforms state-of-the-art methods across most quantitative metrics. Notably, by offloading semantic reasoning to our text-alignment module, SynMind surpasses competing methods based on SDXL while using the much smaller Stable Diffusion 1.4 and a single consumer GPU. Large-scale human evaluations further confirm that SynMind produces reconstructions more consistent with human visual perception. Neurovisualization analyses reveal that SynMind engages broader and more semantically relevant brain regions, mitigating the over-reliance on high-level visual areas.
\end{abstract}


\section{Introduction}

\IEEEPARstart{D}ecoding the internal visual percepts formed in the human brain from visual neural signals is a long-standing and important goal in neuroscience and artificial intelligence. Achieving this goal is crucial for understanding how the human visual system encodes the world we perceive every day and for inspiring the development of more human-like computational vision models. Commonly used neural signals for visual decoding include electroencephalography (EEG) \cite{davis2022brain, jin2024pgcn,kavasidis2017brain2image} and functional magnetic resonance imaging (fMRI) \cite{kamitani2005decoding, miyawaki2008visual,du2025human,du2023decoding}. While EEG offers millisecond-level temporal resolution, it suffers from poor spatial localization. In contrast, fMRI provides whole-brain coverage with high spatial resolution, enabling precise analysis of activity across cortical regions during visual stimulation.

\begin{figure*}
    \centering
    \includegraphics[width=\textwidth]{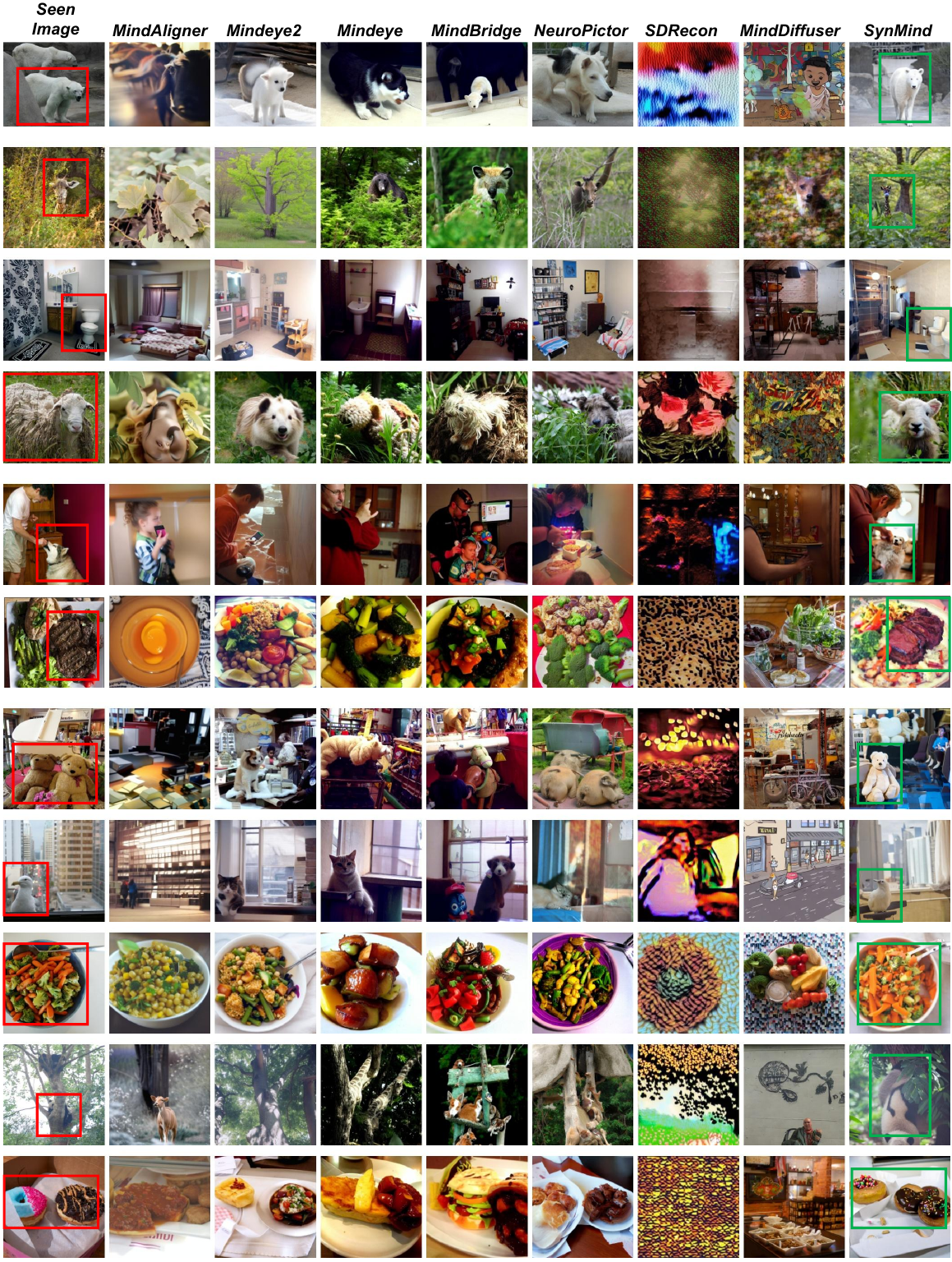}
    \caption{Reconstructed images produced by existing methods with publicly released results. Red boxes mark objects whose semantics are incorrectly reconstructed or misaligned with the original stimuli, while green boxes highlight semantically challenging regions that SynMind is able to reconstruct correctly.}
    \label{fig:new_intro_fig}
\end{figure*}

\IEEEpubidadjcol

Reconstructing visual images from fMRI signals is an intriguing and challenging problem in visual decoding that has evolved significantly. Early work \cite{naselaris2009bayesian, cowen2014neural} relied on traditional machine learning approaches to map fMRI responses to simple grayscale patterns. With the release of large-scale datasets like the Natural Scenes Dataset (NSD) \cite{allen2022massive} and the emergence of latent diffusion models \cite{xu2023versatile, rombach2022high}, recent methods have substantially advanced the fidelity of fMRI-based visual reconstruction. However, a critical issue remains. Although recent methods have improved visual realism, many reconstructed images -- while appearing high-fidelity at first glance -- exhibit significant semantic deviations from the original stimuli, as illustrated in Fig.~\ref{fig:new_intro_fig}. A bowl of vegetable salad might be rendered as vegetable stew; specific objects are often replaced or entirely hallucinated.

This issue largely arises because existing methods predominantly map fMRI signals to implicit, visual-centered intermediate feature spaces—such as CLIP image embeddings \cite{scottimindeye2, scotti2024reconstructing}, depth maps \cite{xia2024dream}, or VAE latents \cite{takagi2023high}. While these representations capture global visual gist effectively, they are often entangled, prioritizing appearance over explicit semantic identity. Consequently, the reconstruction often preserves the ``look'' of the image while misinterpreting the ``meaning'' of its constituents.

In this paper, we aim to preserve visual similarity while significantly improving semantic alignment between the original visual stimuli and the reconstructed images. We argue that the key is to elevate the role of explicit semantic interpretation. Instead of mapping fMRI solely to entangled visual embeddings or short, coarse-grained COCO captions, we leverage language as a high-level scaffold to disentangle these representations. We parse neural activity into comprehensive, sentence-level descriptions that explicit spell out the object categories and spatial organizations of a scene. To achieve this, we leverage multimodal large language models (e.g., Qwen2-VL) to synthesize human-like, multi-granularity textual descriptions. Crucially, to prevent the VLM itself from introducing hallucinations, we ground these descriptions in the original verified annotations (COCO captions), ensuring the semantic signal remains faithful to the ground truth.

Building on this approach, we propose SynMind, a semantic-guided fMRI-to-image reconstruction framework. By explicitly mapping fMRI signals to a rich semantic latent space, we provide strong conditioning to the generative model. This design allows for a more efficient architecture: by offloading the burden of ``figuring out'' the scene semantics to our text-alignment module, SynMind achieves state-of-the-art performance using the computationally lighter Stable Diffusion 1.4, surpassing methods that rely on the much larger SDXL model. Importantly, SynMind is dominated by semantic learning, while the incorporation of visual feature learning serves as an optional auxiliary component.

To comprehensively evaluate the effectiveness of SynMind, we conduct extensive experiments from three complementary perspectives. (i) \textbf{Quantitative and qualitative evaluation.} We adopt the NSD dataset as our testbed. The results demonstrate that SynMind achieves state-of-the-art performance on six out of eight metrics, even when using Stable Diffusion 1.4, while some competing methods rely on the more advanced SDXL model. (ii) \textbf{Human behavioral study.} We conduct a large-scale two-alternative forced-choice (2AFC) human evaluation. Participants consistently prefer SynMind over competing methods, indicating superior perceptual and semantic fidelity. (iii) \textbf{Neuroscience-oriented visualization.} We analyze visual cortical activation patterns associated with SynMind. As semantic representations become progressively finer, we observe increasingly widespread and spatially coherent activation in the visual cortex, in contrast to the more localized activation patterns typically produced by competing methods.

Our contributions are as follows: 

(i) We identify that the reliance on implicit visual embeddings is a primary source of semantic hallucination in fMRI reconstruction and demonstrate the efficacy of explicit semantic disentanglement.

(ii) We introduce MimeVis, a framework that effectively leverages grounded VLMs to synthesize human-like, multi-granularity semantic descriptions of given visual stimuli.

(iii) We propose SynMind, a semantic-guided framework that formulates the core task as learning a projection from fMRI signals to a semantic feature space, while allowing optional integration of visual intermediate latents.

(iv) Extensive experimental results demonstrate that SynMind significantly improves semantic consistency and efficiency, achieving state-of-the-art performance across multiple evaluation settings.

\section{Related Work}

\subsection{Neural Signal for Visual Decoding}
Accurately recording neural activities during external visual stimulation remains a fundamental challenge in both neuroscience research and industry applications. Invasive methods such as spike recordings and calcium imaging  \cite{chen2013ultrasensitive, Orsborn2014closed} offer cellular precision but pose significant surgical and ethical barriers for human use. Consequently, non-invasive approaches dominate human visual decoding research, with EEG \cite{luck2011oxford, Roach2008EventrelatedET} and fMRI \cite{naselaris2009bayesian, miyawaki2008visual} emerging as the most prevalent methods. EEG is primarily used for image classification and identification\cite{sen2020emerging,kumar2019bci,lotte2015electroencephalography} , but its application in image reconstruction is limited due to low spatial resolution. Although fMRI suffers from limited temporal resolution and a relatively low signal-to-noise ratio, it offers a more suitable balance between spatial fidelity and practical feasibility for studying visual representations in humans\cite{gu2022decoding,xia2024dream,scotti2024reconstructing}. 
In this work, we focus on decoding visual stimuli from fMRI signals. Considering the spatial characteristics of fMRI, the availability of large-scale brain–vision paired datasets, and the diversity required to study semantic-level visual representations, we adopt a natural scene dataset as our experimental testbed.

\subsection{Visual Decoding Tasks Based on fMRI}
It is not until recently that decoding natural scene images from fMRI with high relevance becomes realistically possible. Early brain visual decodings aim for much simpler goals such as classification ~\cite{kam2005decoding} and identification~\cite{Horikawa2015GenericDO}, with later gradually extend to recovering low-resolution grayscale visual artifacts, such as 10$\times$10 binary pattern grids ~\cite{naselaris2009bayesian, miyawaki2008visual,yoshida2020natural}. The community remains less active partly because we have not yet had a strong grip on how to generate photorealistic images. The phenomenal success of Variational Autoencoders (VAE) and Generative Adversarial Networks (GANs) allows us to think bigger and pin higher hope on reconstructing finer-grained visual objectives from fMRI, from photorealistic effect to larger resolutions~\cite{gu2022decoding,ozcelik2022reconstruction,seeliger2018generative,fang2020reconstructing,lin2022mind,ren2021reconstructing,shen2019deep}. The trend goes on as latent diffusion models transform the generative image field and so transforms fMRI-based image generation~\cite{xia2024dream, chen2023seeing, lu2023minddiffuser, scottimindeye2, wang2024mindbridge, quan2024psychometry, ozcelik2023brain, huo2024neuropictor}. 
Despite these substantial advances in image resolution and visual quality, a persistent limitation remains: semantic misalignment. Many reconstructed images, while visually sharp and aesthetically convincing, fail to preserve the correct semantic content of the original visual stimuli. Objects may be replaced, altered, or hallucinated, resulting in reconstructions that are visually impressive but semantically incorrect. Addressing this semantic discrepancy is a central motivation of our work.

\subsection{Intermediate Feature Choice in fMRI-to-Image Reconstruction}

In existing fMRI-to-image reconstruction literature, a prevailing assumption is that visual features form the primary representation, while semantic information is treated as auxiliary. To systematically examine this trend, we review a set of representative works published since 2022 \cite{lin2022mind, daimindaligner, scottimindeye2, gong2025mindtuner, lu2023minddiffuser, wang2024mindbridge, scotti2024reconstructing, ozcelik2023brain, xia2024dream, xia2024umbrae, takagi2023high, gu2022decoding, li2025neuraldiffuser}. We choose this time span because the release of Stable Diffusion \cite{rombach2022high} and the NSD dataset \cite{allen2022massive} fundamentally reshaped the field, shifting reconstruction pipelines toward mapping fMRI signals into intermediate feature spaces that serve as conditional inputs for image generation models. The intermediate representations adopted by these methods are summarized in Table~\ref{tab:comparison_methods}. This design choice is partly driven by the limited availability of reliable semantic supervision: existing methods typically rely on COCO captions, which are short, coarse, human-annotated descriptions constrained to salient visual content. As a result, semantic guidance remains weak and underexplored. In contrast, we propose, for the first time, to leverage the visual description capability of vision–language models (VLMs) to synthesize human-like visual semantic descriptions, and use them to more effectively guide the learning of fMRI representations. 

\begin{table}[h]
    \centering
    \caption{Intermediate features used in previous fMRI-to-image reconstruction methods. All existing approaches rely primarily on visual features extracted by different image encoders (e.g., CLIP, SwAV, VD-VAE, etc). While some methods incorporate semantic features into their reconstruction pipelines, these are typically derived solely from COCO-style captions, offering limited semantic richness.}
    \begin{tabular}{lcc}
        \toprule
        \multirow{2}{*}{Methods} & \multicolumn{2}{c|}{Intermediate Feature} \\
        
         & Visual Feature & Semantic Feature  \\
        \midrule
        MindReader\cite{lin2022mind} & CLIP & CLIP(COCO) \\
        Gu\cite{gu2022decoding} & SwAV & - \\
        SDRecon\cite{takagi2023high} & SD-VAE & CLIP(COCO)  \\
        MindDiffuser\cite{lu2023minddiffuser}& VQ-VAE+CLIP & CLIP(COCO)  \\
        BrainDiffuser\cite{ozcelik2023brain} & VD-VAE+CLIP & CLIP(COCO) \\
        MindEye\cite{scotti2024reconstructing} & SD-VAE+CLIP & - \\
        Dream\cite{xia2024dream} & CLIP+Color+MiDas & CLIP(COCO) \\
        UMBRAE\cite{xia2024umbrae} & CLIP & -\\
        MindBridge\cite{wang2024mindbridge} & CLIP & CLIP(COCO)  \\
        MindEye2\cite{scottimindeye2} & SD-VAE+CLIP & -  \\
        NeuroPictor\cite{huo2024neuropictor} & SD-VAE & CLIP(COCO) \\
        MindTuner\cite{gong2025mindtuner} & SD-VAE+CLIP & CLIP(COCO)  \\
        MindAligner\cite{daimindaligner} & CLIP & -\\
        NeuralDiffuser\cite{li2025neuraldiffuser} & SD-VAE+CLIP & CLIP(COCO)\\
        
        \bottomrule
    \end{tabular}
    
    \label{tab:comparison_methods}
\end{table}

\section{Methodology}
\label{sec:method}

In this section, we begin by introducing MimeVis, which aims to replace the otherwise labor-intensive manual semantic annotation process. Inspired by prior prompt-engineering studies~\cite{Kojima2022LargeLM,Wei2022ChainOT,feifei2007perceive}, MimeVis relies on two-round carefully designed prompts that guide the model to simulate human visual perception and generate semantic descriptions at multiple levels of granularity. We then present SynMind, a unified framework for fMRI-to-image reconstruction. SynMind is composed of four core modules: a subject-wise mapper, a subject-shared semantic encoder, a subject-shared visual encoder, and a Semantic-Aware Renderer, which together enable effective semantic alignment and high-quality visual reconstruction. Overall, SynMind represents a concrete instantiation of integrating the human-like visual semantic description capabilities of vision–language models into the fMRI-to-image reconstruction paradigm.

\begin{figure*}[htbp]
    \centering
    \includegraphics[width=0.98\textwidth]{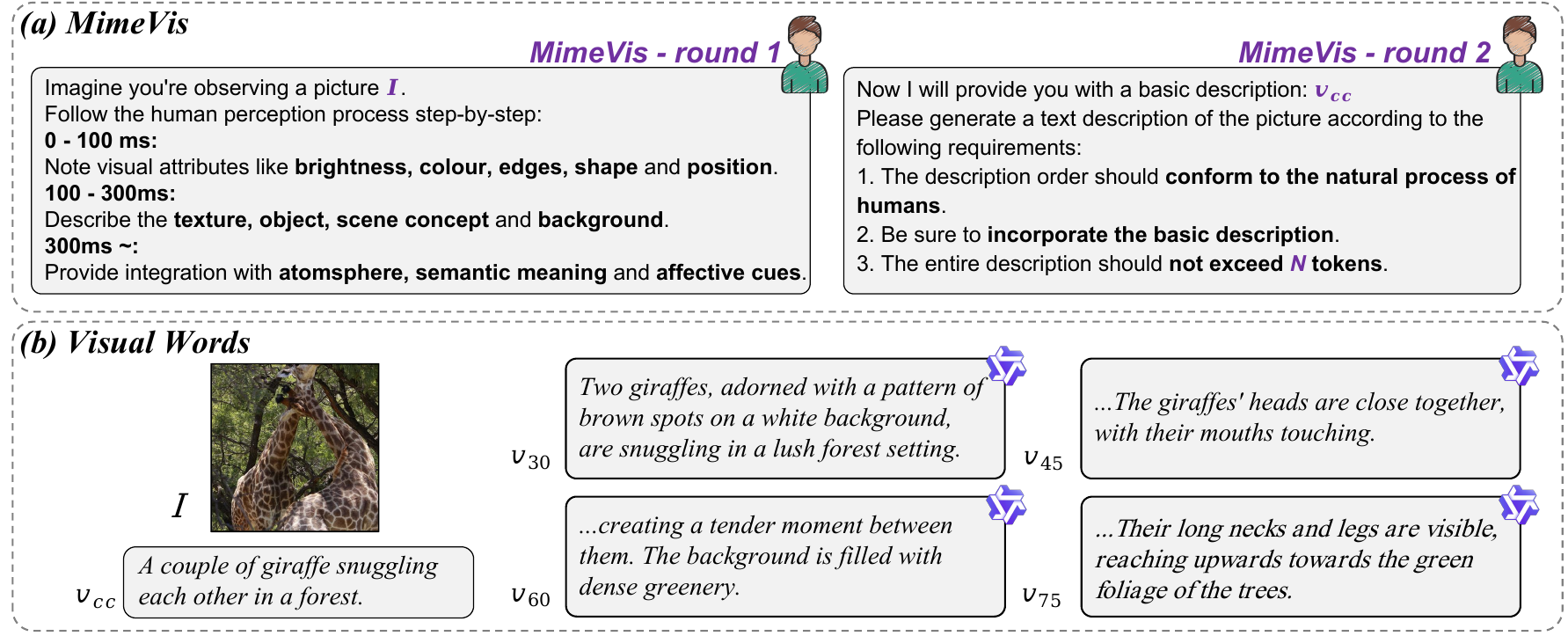}
    \caption{(a) Illustration of the two-round pipeline of \textit{MimeVis} to prompt \textit{Qwen2-VL} generate textural descriptions of an image while maximally simulating human visual perception process. In the first round, we give \textit{Qwen2-VL} an open mic and encourage it to cover as many aspects as it could of an image. In the second round, we start to enforce constraints on both the lengths and contents of a finalized caption and importantly introduce COCO caption as a ground truth description to reduce synthetic hallucinations. (b) An example of the results from \textit{MimeVis}. $I$ and $v_{cc}$ are visual stimuli and its original COCO caption, while $v_{N}$ represents outputs of \textit{MimeVis} under different lengths limits $N$.}
    \label{fig:framework}
\end{figure*}

\subsection{MimeVis}

To ensure we have a reasonable expectation of what information within an image can be possibly abstracted in fMRI, we need to look into how fMRI data is collected. We choose Natural Scene Dataset (NSD) ~\cite{allen2022massive} which is a large-scale fMRI dataset conducted at ultra-high-field (7T) strength and contains fMRI recordings across eight healthy adult subjects. Each subject is required to view 9,000$\sim$10,000 natural scene visual stimuli curated from COCO where the whole process is divided into 30$\sim$40 independent sessions. Each session includes 750 trials. At each trial, the subject is shown visual stimuli and asked to memorize as many details as possible within three seconds. The spontaneous mental activity is then recorded using whole-brain gradient-echo EPI at an isotropic resolution of 1.8 mm and a repetition time (TR) of 1.6s on a 7T scanner and stored in the form of a fMRI. 

Now that we know fMRI only corresponds to an image viewing experience with a span of three seconds, it's important that our synthetic caption recognizes this reality and refrains from generating either too sketchy or too exhaustive descriptions. We also seek reference to the neuroimaging community \cite{brady2008visual,hollingworth2001see} to reckon if there is any finding that can help us understand what aspects of an image are perceived during this short time -- we know human visual system initiates complex processing procedures in just a few hundred milliseconds, where in the first 100 millisecond, low-level visual attributes such as brightness, color, edge, shape and position are extracted. After this initial phase, object and scene concepts occur, followed by the emergence of background understandings. Finally, vision delves deeper by interpreting things more intimate, including making statements on the connections between certain visual entities and personal feelings; for example how a specific flower and its scent could cause the recall of a memory.

To this end, we propose MimeVis which prompts an open-source multi-modal foundation model to generate captions for visual stimuli that hopefully cover what people actually see featuring both low-level and high-level abstractions. Specifically, we employ two-round interactive prompting on $\Phi_{VLM}$ - \textit{Qwen2-VL} as illustrated in Fig.~\ref{fig:framework}(a). In the first round, given a visual stimulus $I$, we ask $\Phi_{VLM}$ to generate a complete set of visual descriptions including basic attributes like brightness color, texture but also affective cues and semantic objects. In the second round, we ask $\Phi_{VLM}$ to use the output from previous as context and reorganize its autoregressive caption generations with two important constraints: i) to maximally avoid AI hallucinations, we use COCO caption $V_{cc}$ that comes inherent to $I$ (because $I$ is from a curated sample collection of COCO) and explicitly command whatever $\Phi_{VLM}$ generates must comply with the basic meanings encoded in $V_{cc}$. ii) to specify maximum number of $N$ visual words that $\Phi_{VLM}$ can generate. This not only enforces $\Phi_{VLM}$ to output precise textual descriptions but also make it more feasible as an input to text-to-image model later which typically has a limit on the length of input text words. The whole process can be formulated as:
\begin{equation}
    V_{N} = \Phi_{VLM}(\Phi_{VLM}(I), V_{cc}, N).
\end{equation}

\noindent Overall, we associate each $I$ with visual words across five abstraction levels, ranging from coarse to fine: $V = \{v_{cc}, v_{30}, v_{45}, v_{60}, v_{75}\}$ (Fig.~\ref{fig:framework}(b)). In the following section, we denote the fMRI signal as $x$ and its corresponding visual stimulus as $I$, which is annotated by synthetic semantic captions $V$ (for sake of cleanness, we use $V$ to represent a synthetic caption at any abstraction level without loss of clarity). Additionally, since $I$ is curated from COCO dataset, we have a complete list of object categories that presents in each $I$. We represent the objects in $I$ by the set $O = \{o_{1}, o_{2}, \dots, o_{C}\}$, where $C$ denotes the number of object categories.

\begin{figure*}[htbp]
    \centering
    \includegraphics[width=0.98\textwidth]{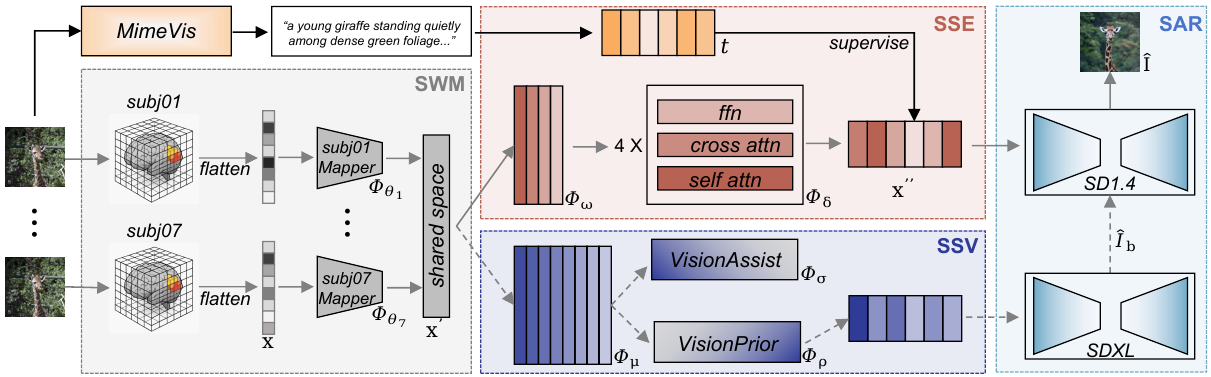}
    \caption{Overview of the proposed fMRI-to-image reconstruction framework, \textbf{SynMind}. The framework consists of four core modules: the Subject-Wise Mapper (SWM), the Subject-Shared Semantic Encoder (SSE), the Subject-Shared Visual Encoder (SSV), and the Semantic-Aware Renderer (SAR). In the figure, black dashed arrows ($\dashrightarrow$) indicate operations used only during training, while gray dashed arrows (\textcolor{gray}{$\dashrightarrow$}) denote optional components that can be enabled or disabled at inference time. Solid gray arrows (\textcolor{gray}{$\rightarrow$}) illustrate the essential semantic-guided workflow from fMRI signals to image reconstruction. }
    \label{fig:synmind}
\end{figure*}

\subsection{SynMind}
\label{sec:annotation}

\label{sec:framework}
SynMind is composed of four core modules: a Subject-Wise Mapper (SWM), a Subject-Shared Semantic Encoder (SSE), a Subject-Shared Visual Encoder (SSV), and a Semantic-Aware Renderer (SAR) (Fig.~\ref{fig:synmind}). The SWM first projects fMRI signals of varying dimensionality from different subjects into a unified shared latent space. Based on this shared representation, the SSE and SSV further map the unified fMRI latents into well-structured semantic and visual intermediate feature spaces, respectively. Finally, SAR conditions on these mapped fMRI features to render the final results.

\noindent \textbf{\textit{Subject-Wise Mapper}}: We denote the subject-specific linear layer $\Phi_{\theta_k}$ with trainable parameters $\theta_k$ to project $x$ into a predefined dimension $d$, such as $x' = \Phi_{\theta_k}(x) \in \mathbb R^d$. Furthermore, to ensure that $x'$ retains essential cross-subject shared information, we employ an auxiliary classifier that takes $x'$ as input and performs multi-class classification to verify whether $x'$ can be accurately assigned to its corresponding multi-object categories. The training objective likes:

\begin{equation}
    \mathcal{L}_{\mathrm{SWM}} 
    = - \sum_{c=1}^{M} \Bigl[ 
        y_{c} \log\bigl(\hat{y}_{c}\bigr) 
        + \bigl(1 - y_{c}\bigr) \log\bigl(1 - \hat{y}_{c}\bigr) 
    \Bigr],
\end{equation}

where $y_{c}$ is the binary label indicating the presence of object $c$ in $I$, $\hat{y}_{c}$ is the predicted label based on $x'$, and $M$ denotes the total number of object categories in the COCO dataset. 

\noindent\textbf{\textit{Subject-Shared Semantic Encoder:}} 
After projecting the subject-specific fMRI signals into a shared fixed-length vector space, the next step is to establish a connection that aligns it with pretrained well-structured semantic feature space. SSE operates in two phases: encoding and denoising. For the encoding phase, we utilize a four-layer MLP $\Phi_\omega$ with residual blocks in the first three layers, each of dimension $d$, and a final layer that maps from $d$ to $N \times D$, where $D$ is the embedding dimension of pretrained semantic feature space. To construct a robust and general latent space, we employ the \textit{SoftCLIP} loss to distill knowledge from a teacher model:
\begin{align}
\mathcal{L}_\text{SoftCLIP} 
  &= - \sum_{j=1}^B 
  \Biggl[
  \frac{\exp\Bigl(\frac{t \cdot t_j}{\tau}\Bigr)}
       {\sum_{m=1}^B \exp\Bigl(\frac{t \cdot t_m}{\tau}\Bigr)}
  \notag\\
  &\qquad \times
  \log\Bigl(
       \frac{\exp\Bigl(\frac{\Phi_\omega(x') \cdot t_j}{\tau}\Bigr)} 
            {\sum_{m=1}^B \exp\Bigl(\frac{\Phi_\omega(x') \cdot t_m}{\tau}\Bigr)}
  \Bigr)
  \Biggr],
\end{align}
where $B$ is the batch size, $t$ is the target semantic latent of $x'$, $t_j$ is the $j$-th target semantic latent corresponding to the $x'$-encoded batch, and $\tau$ is the temperature parameter of the softmax function. Given the limited availability of fMRI-image pairs, we further deploy a commonly used data augmentation method - MixCo~\cite{kim2020mixco} to help large models in low-data regimes, as formulated by\footnote{Since MixCo is merely a data augmentation technique that does not alter the latent dimension or meaning, we continue to use $x'$ to represent the augmented latent for clarity and ease of reading.}

\begin{equation}
    x_{mix_{i,j}} = \alpha x'_i + (1 - \alpha) x'_j,
\end{equation}

\noindent where $\alpha \sim \text{Beta}(1,1)$ is a mixing coefficient sampled from the Beta distribution, and $x'_j$ is the $j$-th latent representation to the $x'$-encoded batch. After encoding, the resulting embeddings $\Phi_\omega(x') \in \mathbb R^{N\times D}$ condition a diffusion prior $\Phi_\delta$, yielding a denoised set of fMRI semantic embeddings $x''$:
\begin{equation}
x'' = \Phi_\delta\bigl(\Phi_\omega(x')\bigr) \;\in\; \mathbb{R}^{N\times D}.
\end{equation}
Both $\Phi_\omega$ and $\Phi_\delta$ are trained under an MSE loss to shrink the distance between the predicted $x''$ and target semantic features $t$:
\begin{align}
    \mathcal{L}_\text{MSE} 
    = \bigl\| t - \Phi_\omega(x') \bigr\|^2 
    \;+\; \lambda_{denoising}\bigl\| t - x'' \bigr\|^2.
\end{align}

\noindent\textbf{\textit{Subject-Shared Visual Encoder:}} The objective of the Subject-Shared Visual Encoder (SSV) is to map the shared fMRI embedding $x'$ into an intermediate visual feature space. To this end, SSV is composed of three sequential components that progressively enhance the visual fidelity of the fMRI representation. First, an eight-layer MLP $\Phi_{\mu}$ encodes the shared fMRI embedding $x'$ into a visual-aware representation $x'_v$. In parallel, a VisionAssist module $\Phi_{\sigma}$ provides auxiliary supervision signals to guide the effective learning of $x'_v$. Specifically, $\Phi_{\mu}$ consists of eight fully connected layers with hidden dimensions of $[4096, 4096]$. The VisionAssist module introduces two complementary supervision objectives: (i) $x'_v$ is upsampled via a CNN to a spatial resolution of $[64, 64, 4]$ and aligned with the VAE latent of the original visual stimulus using an $\ell_1$ loss; (ii) $x'_v$ is further processed by three CNN layers to obtain a feature map of size $[512, 7 \times 7]$, which is aligned with the ConvNeXt-XXL embedding of the original stimulus using a SoftCLIP loss.  Next, a VisionPrior module $\Phi_{\rho}$ further refines the visual-aware representation. It first reshapes $x'_v$ into a latent of size $[256 \times 1664]$ and then applies a causal transformer to denoise and enhance the representation, producing a refined visual fMRI embedding $x''_v$ that is closer to the CLIP image embedding of the corresponding visual stimulus.

Therefore, the overall training objective of SSV is defined as\footnote{The detail implementation and training objectives of SSV can be found in supplementary.}:

\begin{equation}
    \mathcal{L}_{\text{SSV}} = \lambda_{prior}\mathcal{L}_{prior} + \lambda_{assist}\mathcal{L}_{assist}.
\end{equation}

\noindent\textbf{\textit{Semantic-Aware Render:}} During inference, the refined visual fMRI embedding $x''_v$ produced by the VisionPrior module in SSV is first fed into a pretrained SDXL UnCLIP model to generate an initial coarse reconstruction $\hat{I}_b$, which is used to initialize the variational autoencoder (VAE) latent. In parallel, the output of the Subject-Shared Semantic Encoder (SSE), denoted as $x''$, serves as the semantic conditioning signal. The initialized VAE latent $\hat{I}_b$ and the semantic embedding $x''$ are then jointly input into Stable Diffusion 1.4. At each denoising step, $x''$ interacts with the latent representation through cross-attention, guiding the diffusion process toward semantically coherent structures. Through this iterative refinement, Stable Diffusion synthesizes the final reconstructed image $\hat{I}$.  

Notably, the initialization of the VAE latent using the visual embedding $x''_v$ is optional. The inference process can be performed using only the semantic embedding $x''$ as conditioning, without any visual initialization. This flexible design enables a controlled investigation of the role of semantic information as an intermediate representation in fMRI-to-image reconstruction, allowing us to explicitly assess its contribution independent of visual priors.

\section{Experiments}
\label{sec:exp}

\subsection{Implementation}
\textbf{Dataset}: We use the Natural Scenes Dataset (NSD) \cite{allen2022massive}, the largest publicly available set, widely used for fMRI to image reconstruction. We use the ``nsdgeneral'' ROI to represent brain activity, as it encompasses voxels with the highest activation during visual stimulus presentation. Following the standard train/test split employed in prior NSD reconstruction studies~\cite{lin2022mind, xia2024dream, ozcelik2023brain}, we use the shared set of 982 visual stimuli as the test set, presented to all subjects. For the test set, we average over multiple repetitions of each image, while the training set is not averaged.

\noindent \textbf{Architecture and Training Details}: The SWM is implemented as a linear layer for each subject, mapping $\mathbb{R}^{Z_k} \rightarrow \mathbb{R}^{4096}$, where $Z_k$ represents the number of available voxels in the ``nsdgeneral'' ROI for subject $k$. For $\mathcal{L}_{\text{SWM}}$, the model categorizes into 80 object categories. The $SSE$ encoding module is composed of four residual linear layers with output channels specified as $\{4096, 4096, 4096, N \times 768\}$\footnote{To accelerate training and convergence, $N*768$ is uniformly padded to $77*768$.}. And a Transformer architecture with 4 layers is followed, each containing 8 attention heads of dimension 64 to output refined semantic feature. SynMind is trained using the Adam optimizer with a CosineAnnealingLR learning rate scheduler. The loss weights are set to $\lambda_{\text{SoftCLIP}} = 0.1$, $\lambda_{\text{SWM}} = 0.05$, $\lambda_{\text{denoising}} = 0.5$, $\lambda_{\text{prior}} = 1$, and $\lambda_{\text{assist}} = 0.33$. The semantic-only variant, SynMind*, is trained under the same hyperparameter configuration and is implemented on a single NVIDIA RTX 3090 GPU.

\noindent \textbf{Evaluation Metric}: To comprehensively assess the quality of the reconstructed images, we follow previous works\cite{lin2022mind,gu2022decoding,takagi2023high,scotti2024reconstructing,xia2024dream} to evaluate both low-level and high-level consistency.
On the low-level aspect, we evaluate the pixel-wise correspondence (PixCorr), structural integrity (SSIM), AlexNet(2) (Alex(2)), and AlexNet(5) (Alex(5)) between the reconstructed images and visual stimuli to measure low-level fidelity.
To evaluate the preservation of high-level semantic content, we extracted features using specific networks, including InceptionV3 (Inception), CLIP-L (CLIP), EfficientNet-B1 (EffNet-B) and SwAV-ResNet50 (SwAV). 
For Alex(2), Alex(5), Inception and CLIP, we assess feature-level similarity through a two-way identification task.

\subsection{Quantitative Results}

In Table~\ref{tab:main_quantitative}, we present the average quantitative results across four subjects (01, 02, 05 and 07), following the same settings as other methods. SynMind and its semantic-only variant SynMind* consistently outperform state-of-the-art methods across high-level evaluation metrics. In particular, they achieve nearly a 2\% improvement on the Inception and CLIP scores, demonstrating that incorporating fine-grained semantic information enables more accurate decoding of semantic perception from fMRI signals, even in the absence of visual cues (SynMind*). Moreover, SynMind surpasses 14 prior methods on six out of the eight reported metrics. This improvement can be attributed to the inherent difficulty of directly decoding pixel-level visual details from fMRI signals; instead, targeting the decoding of abstract fine-grained semantic representations, complemented by pixel-level visual cues, provides a more effective pathway for enhancing reconstruction quality.

\subsection{Qualitative Results}

\begin{figure*}[htbp]
    \centering
    \includegraphics[width=\linewidth]{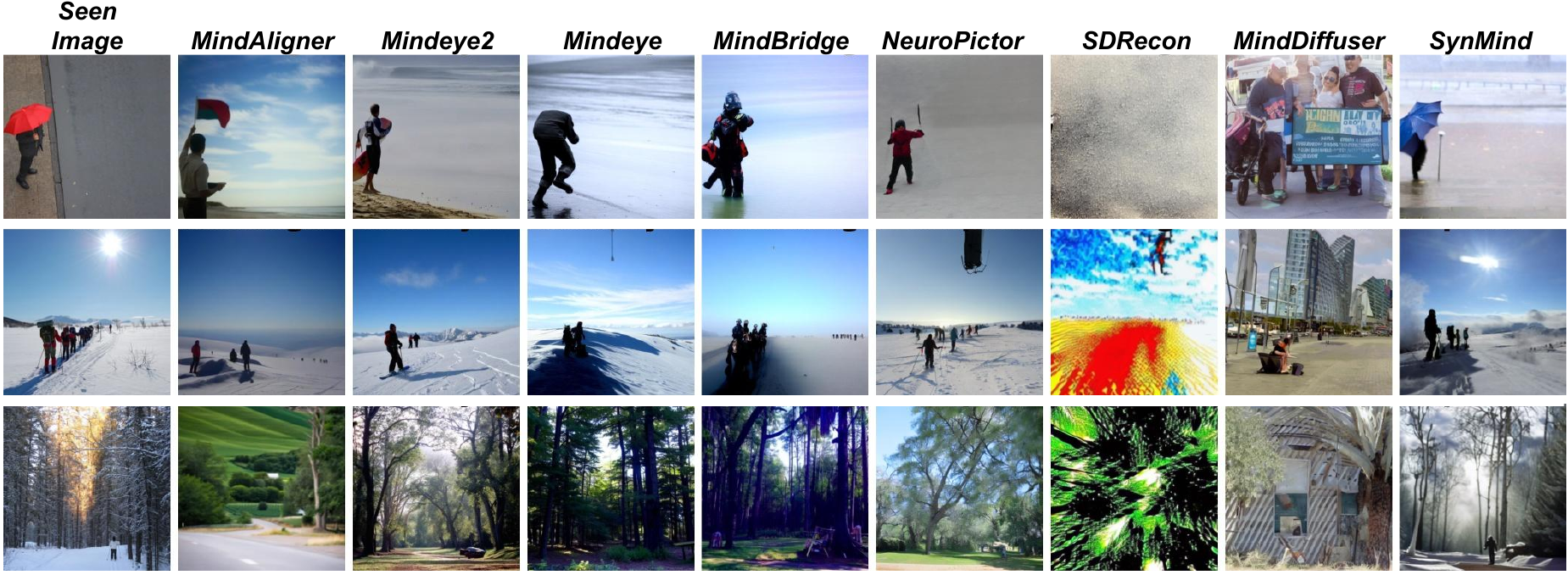}
    \caption{Qualitative comparison of fMRI-based image reconstructions between SynMind and existing state-of-the-art methods. Towards fair comparison, whenever possible, we directly replicated their reconstruction results from their published papers. In cases where such results are not available, we regenerate the results using their publicly released model weights.}
    \label{fig:main_qualitative}
\end{figure*}

We show the main qualitative results in Fig.~\ref{fig:new_intro_fig} and Fig.~\ref{fig:main_qualitative}. In Fig.~\ref{fig:new_intro_fig}, we present qualitative results in which objects that are ignored or incorrectly reconstructed by prior methods are explicitly highlighted, while the same objects are correctly recovered by SynMind. These examples provide an intuitive demonstration of SynMind’s clear advantage in preserving semantic fidelity.

And we can find more cues of the superiority of SynMind in Fig.~\ref{fig:main_qualitative}. For salient foreground objects, as illustrated in the first row of Fig.~\ref{fig:main_qualitative}, only SynMind successfully reconstructs the scene of \emph{``a person under an umbrella''}, whereas other methods fail to capture the correct semantic content. For background semantics, the second row of Fig.~\ref{fig:main_qualitative} shows that SynMind is the only method that correctly identifies and reconstructs the \emph{sun} in the sky. In contrast, competing approaches either overemphasize generic background attributes such as weather conditions or hallucinate non-existent objects, as observed in the results produced by MindDiffuser. Finally, for small but crucial objects, the third row of Fig.~\ref{fig:main_qualitative} demonstrates that SynMind uniquely captures the \emph{person} located in the lower-right corner of the image. Other methods tend to focus predominantly on the large-area \emph{forest} region and overlook this critical semantic detail. Together, these results highlight SynMind’s strong capability to maintain fine-grained semantic alignment across a wide range of visual contexts.

\begin{table}[h]
\caption{2AFC human study results. The Rate denotes the percentage of trials in which SynMind (or SynMind*, depending on the comparison) was preferred over the competing method, with the chance level set to 50\%. The $\boldsymbol{\Delta}$ column reports the margin above the chance level. SynMind represents the complete framework, while SynMind* denotes the variant without visual guidance, relying solely on semantic guidance.}
\centering
\small
\setlength{\tabcolsep}{6pt}
\renewcommand{\arraystretch}{1.2}
\begin{tabular}{lccc}
\toprule
\textbf{Comparison} & \textbf{Preferred} & \textbf{Rate} & $\boldsymbol{\Delta}$($\%$) \\
\midrule
SynMind~ vs SynMind*     & SynMind & 60.44\% & +10.44 \\
SynMind~ vs MindEye2     & SynMind & 56.44\% &  +6.44 \\
SynMind* vs MindDiffuser & SynMind* & 97.21\% & +47.21 \\
SynMind* vs MindBridge   & SynMind* & 72.98\% & +22.98 \\
SynMind* vs MindEye      & SynMind* & 61.02\% & +11.02 \\
SynMind* vs NeuroPictor  & SynMind* & 53.03\% &  +3.03 \\
SynMind* vs MindEye2     & SynMind* & 54.23\% &  +4.23 \\
SynMind* vs Takagi       & SynMind* & 98.82\% & +48.82 \\
SynMind* vs MindAligner  & SynMind* & 88.32\% & +38.32 \\
\bottomrule
\end{tabular}
\label{tab:human_study}
\end{table}

\begin{table*}[htbp]
    \centering
    \caption{Quantitative comparisons between SynMind and existing fMRI-to-image reconstruction methods. Results are averaged across subjects 01, 02, 05, and 07 from the NSD Dataset. $\#$ indicates the results are reported under 1 hour data setting.}
    \renewcommand{\arraystretch}{1.2}
	\resizebox{\textwidth}{!}{
		\begin{tabular}{l|cccc|cccc}
			\toprule
			\multirow{2}{*}{Method}  & \multicolumn{4}{c|}{Low-Level} & \multicolumn{4}{c}{High-Level}\\
            
			~  & PixCorr $\uparrow$ & SSIM $\uparrow$ & AlexNet(2) $\uparrow$ & AlexNet(5) $\uparrow$ & Inception $\uparrow$ & CLIP $\uparrow$  & EffNet-B $\downarrow$ & SwAV $\downarrow$\\
			\midrule
		      MindReader~\cite{lin2022mind}(NeurIPS'22)  &- &- &- &-  & 78.2\% &- &- &-\\
            Gu\cite{gu2022decoding}(MIDL'23)  & .325 & .150 & - & - & - & - & .862 & .465 \\
            
			SDRecon~\cite{takagi2023high}(CVPR'23)  &- &- &  83.0\% & 83.0\%  & 76.0\% & 77.0\% & - & -\\
			MindEye~\cite{scotti2024reconstructing}(NeurIPS'24) &  .309 & .323 &   94.7\% &  97.8\% &  93.8\% &  94.1\% &  .645 &  .367 \\
            DREAM~\cite{xia2024dream}(WACV'24)  & .274   & .328 & 93.9\%      &    96.7\%      &   93.4\%     &  94.1\% &  .645     &    .418 \\
            UMBRAE\cite{xia2024umbrae}(ECCV'24) &  .283    &   .341 &  95.5\%     &   97.0\%     &    91.7\%     &   93.5\% &   .700     &   .393 \\
            MindBridge\cite{wang2024mindbridge}(CVPR'24)  & .151 & .263 & 87.7\% & 95.5\% & 92.4\% & 94.7\% & .712 & .418 \\
            NeuralDiffuser\cite{li2025neuraldiffuser}(TIP'25)  & .318 & .299 & 95.3\% & 98.3\% & 95.4\% & 95.4\% & .663 & .401 \\
            MindDiffuser\cite{lu2023minddiffuser}(ACMMM'23)  & .354 & .278 & - & - & - & 76.5\% & - & - \\
            BrainDiffuser\cite{ozcelik2023natural}(SciRep'23)  & \textbf{.356} & .254 & 94.2\% & 96.2\% & 87.2\% & 91.5\% & .775 & .423 \\
            MindEye2\cite{scottimindeye2}(ICML'24)  & .322 & \textbf{.431} & 96.1\& & 98.6\% & 95.4\% & 93.0\% & .619 & .344 \\
            MindTuner\cite{gong2025mindtuner}(AAAI'25)  & .322 & .421 & 95.8\% & 98.8\% & 95.6\% & 93.8\% & .612 & \textbf{.340} \\
            MindAligner$\#$\cite{daimindaligner}(ICML'25)  & .206 & .414 & 85.6\% & 91.3\%  & 83.0\%  & 81.2\% & .802  & .463 \\
            NeuroPictor\cite{huo2024neuropictor}(ECCV'24)  & .229 & .375 & \textbf{96.5\%} & 98.4\% & 94.5\% & 93.3\% & .639 & .350 \\
            \cdashline{1-9}
          
            SynMind &  .285 & .407 & \textbf{96.5\%} & \textbf{98.8\% }& \textbf{97.8\%} & \textbf{96.9\%} & \textbf{.570} & \textbf{.340} \\
             SynMind*  &  .096& .309& 77.9\% & 89.9\% & \textbf{97.2\%} & \textbf{96.9\%} &  \textbf{.609}&  .413\\
            
             \bottomrule
		\end{tabular}
	}
    
    \label{tab:main_quantitative}
\end{table*}

\subsection{Human Study}
The quantitative evaluation metrics objectively measure the similarity between the reconstructed images and the original visual stimuli from both low-level and high-level perspectives. Beyond these objective metrics, we further conducted a human subjective evaluation to examine whether the reconstructed images faithfully reflect what participants actually perceived. To ensure reliability, the human study was designed as a two-alternative forced choice (2AFC) test: participants were first shown the original visual stimulus for three seconds (following the fMRI data collection protocol of NSD), and were then presented with two reconstructed images generated by different methods. They were asked to choose ``which image is more similar to the one you just saw''. To verify the attentiveness and reliability of participants' responses, a sentinel trial was inserted every 20 trials, in which one of the two options was the original stimulus. If a participant failed to select the original image in these trials, all of their responses were excluded from the analysis. To avoid bias, the order of the two reconstruction methods was randomized across trials. 

For a fair comparison, we include seven reconstruction methods: Takagi~\cite{takagi2023high}, MindDiffuser~\cite{lu2023minddiffuser}, MindBridge~\cite{wang2024mindbridge}, MindEye~\cite{scotti2024reconstructing}, NeuroPictor~\cite{huo2024neuropictor}, MindEye2~\cite{scottimindeye2}, and MindAligner~\cite{daimindaligner}. All of these methods have publicly released either their source code with pretrained model weights, enabling us to reproduce their results, or directly provided the reconstructed images on the test set for evaluation.

Table~\ref{tab:human_study} presents the quantitative results of the human study. As shown, except for MindEye2 and SynMind, SynMind* achieves the highest selection rate among all compared methods—regardless of whether they rely on visual, semantic, or joint information. This demonstrates that decoding fine-grained semantic representations from fMRI signals alone can already yield satisfactory perceptual results. Furthermore, when incorporating visual information as an additional decoding target, SynMind surpasses MindEye2, even though MindEye2 employs a finetuned SDXL model, whereas SynMind uses the pretrained Stable Diffusion 1.4 backbone (only 25\% of SDXL’s parameters) without any additional finetuning.  These results collectively suggest that while recent diffusion-based methods have significantly improved visual realism, explicit semantic alignment—as emphasized in SynMind—plays a critical role in producing reconstructions that are not only visually plausible but also semantically faithful to original stimuli as judged by human.

\begin{figure*}[t]
    \centering
    \caption{Visualization of ROI importance for Subjects~01, 02, 05, and~07, derived from the learned weights of the first layer in the Subject-Wise Mapper (SWM) of SynMind*. For each subject, voxel-wise weights are first averaged across all modules and then normalized such that the total activation weight across voxels sums to 1, where red and yellow correspond to lower and higher importance, respectively. $N=\{30,45,60,75\}$ indicates different levels of granularity of the intermediate semantic features extracted by MimeVis. White text labels denote ROI names and are displayed at fixed reference locations for visualization clarity rather than exact subject-specific ROI centroids.}
    \includegraphics[width=\linewidth]{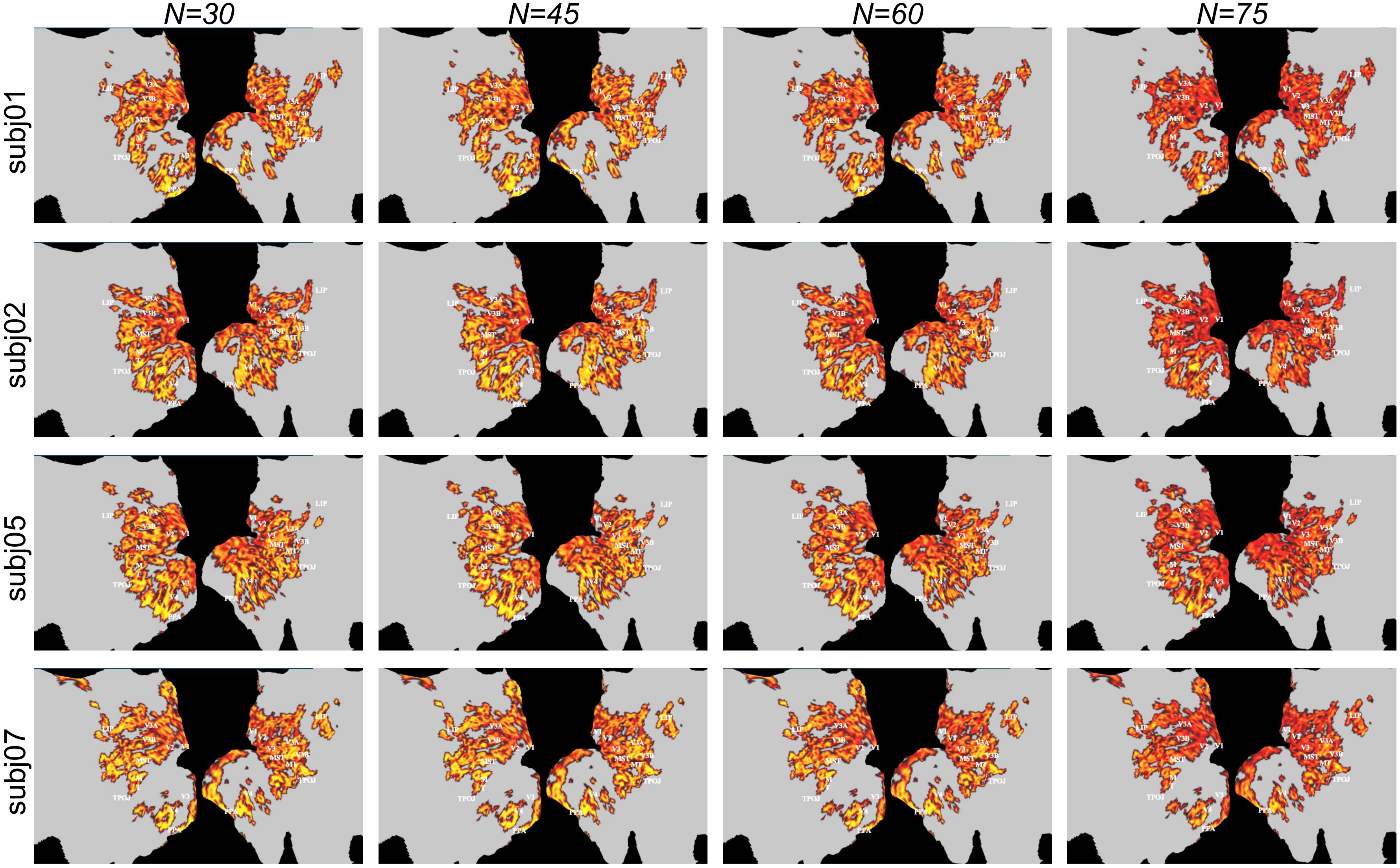}

    \label{fig:neuro}
\end{figure*}

\subsection{NeuroScience Interpretability}

To elucidate how semantic granularity influences cortical engagement during fMRI feature learning, we examine the activation weight distributions across different semantic levels, as shown in Fig.~\ref{fig:neuro}.
At the coarse semantic level ($N=30$), activations are predominantly localized in higher-order visual and associative cortices, including the PPA, TPOJ, etc. These regions exhibit bright yellow responses, indicating stronger activation weights and higher decoding relevance. In contrast, early visual cortices (V1–V3) display weaker, reddish activation, suggesting that coarse-grained semantic reconstruction relies mainly on abstract and scene-level representations, with limited contribution from low-level visual features. As the semantic granularity becomes finer ($V_{75}$), activation patterns expand and redistribute across the visual hierarchy. The early and mid-level visual areas (V1–V4) exhibit increased activation strength and enhanced spatial continuity, indicating a re-engagement of low-level perceptual representations. Meanwhile, high-level associative regions such as PPA and TPOJ remain active but lose their relative dominance, reflecting a more balanced contribution between hierarchical cortical areas. Taken together, these findings reveal a clear shift in cortical decoding dynamics—from a top-down, high-level dominance under coarse semantics toward an integration of high- and low-level visual areas under fine semantics.

\section{Conclusion}
\label{sec:conclusion}

This work revisits the long-standing challenge of reconstructing visual perception from fMRI signals and identifies semantic hallucination as a core bottleneck that persists despite recent advances in generative modeling. While prior methods have achieved impressive photo-realism, they often fail to preserve the semantics of salient objects, leading to reconstructions that are visually plausible yet conceptually incorrect. We argue that this limitation stems from an underestimation of the role of semantic interpretation in fMRI decoding.
To address this issue, we propose SynMind, which elevates semantic understanding to a first-class component of the reconstruction pipeline. By parsing fMRI signals into rich, sentence-level descriptions that reflect the chain of human thoughts elicited by visual stimuli, SynMind shifts the emphasis from coarse, short-form semantics to comprehensive, multi-granularity semantic representations. Leveraging multimodal large language models to synthesize human-like descriptions, SynMind strengthens semantic constraints in downstream image generation while remaining compatible with standard diffusion-based frameworks.
Extensive experiments demonstrate that this semantic-centric design yields consistent and meaningful gains. SynMind outperforms significantly larger and more resource-intensive methods across most quantitative metrics, is strongly preferred by human evaluators in large-scale perceptual studies, and exhibits more balanced neural engagement across brain regions. Notably, these improvements are achieved without direct visual supervision, underscoring the effectiveness of semantic guidance alone in aligning reconstructed images with human perception.
Overall, SynMind provides compelling evidence that advancing fMRI-to-image reconstruction requires not only more powerful generative models, but a deeper integration of semantic cognition. By re-centering semantics in the decoding process, this work offers a principled and efficient path toward reconstructions that are not just visually convincing, but semantically faithful to what the brain truly perceives.

\newpage
\bibliographystyle{IEEEtran}
\bibliography{refer}

@String(CVPR= {IEEE Conf. Comput. Vis. Pattern Recog.})

@String(ICCV= {Int. Conf. Comput. Vis.})

@String(ECCV= {Eur. Conf. Comput. Vis.})

@String(ACMMM= {ACM Int. Conf. Multimedia})

@String(AAAI = {AAAI})

@String(CVPR  = {CVPR})

@String(ICCV  = {ICCV})

@String(ECCV  = {ECCV})

@String(ACMMM = {ACM MM})

@article{allen2022massive,
  title={A massive 7T fMRI dataset to bridge cognitive neuroscience and artificial intelligence},
  author={Allen, Emily J and St-Yves, Ghislain and Wu, Yihan and Breedlove, Jesse L and Prince, Jacob S and Dowdle, Logan T and Nau, Matthias and Caron, Brad and Pestilli, Franco and Charest, Ian and others},
  journal={Nature neuroscience},
  volume={25},
  pages={116--126},
  year={2022}
}

@inproceedings{lin2022mind,
  title={Mind reader: Reconstructing complex images from brain activities},
  author={Lin, Sikun and Sprague, Thomas and Singh, Ambuj K},
  booktitle={NeurIPS},
  year={2022}
}

@inproceedings{takagi2023high,
  title={High-resolution image reconstruction with latent diffusion models from human brain activity},
  author={Takagi, Yu and Nishimoto, Shinji},
  booktitle={CVPR},
  year={2023}
}

@inproceedings{scotti2024reconstructing,
  title={Reconstructing the mind's eye: fMRI-to-image with contrastive learning and diffusion priors},
  author={Scotti, Paul and Banerjee, Atmadeep and Goode, Jimmie and Shabalin, Stepan and Nguyen, Alex and Dempster, Aidan and Verlinde, Nathalie and Yundler, Elad and Weisberg, David and Norman, Kenneth and others},
  booktitle={NeurIPS},
  year={2024}
}

@inproceedings{scottimindeye2,
  title={MindEye2: Shared-Subject Models Enable fMRI-To-Image With 1 Hour of Data},
  author={Scotti, Paul Steven and Tripathy, Mihir and Torrico, Cesar and Kneeland, Reese and Chen, Tong and Narang, Ashutosh and Santhirasegaran, Charan and Xu, Jonathan and Naselaris, Thomas and Norman, Kenneth A and others},
  booktitle={ICML},
  year={2024}
}

@inproceedings{xia2024dream,
  title={Dream: Visual decoding from reversing human visual system},
  author={Xia, Weihao and de Charette, Raoul and Oztireli, Cengiz and Xue, Jing-Hao},
  booktitle={WACV},
  year={2024}
}

@inproceedings{xia2024umbrae,
  title={Umbrae: Unified multimodal brain decoding},
  author={Xia, Weihao and de Charette, Raoul and Oztireli, Cengiz and Xue, Jing-Hao},
  booktitle={ECCV},
  year={2024},
}

@inproceedings{wang2024mindbridge,
  title={Mindbridge: A cross-subject brain decoding framework},
  author={Wang, Shizun and Liu, Songhua and Tan, Zhenxiong and Wang, Xinchao},
  booktitle={CVPR},
  year={2024}
}

@article{naselaris2009bayesian,
  title={Bayesian reconstruction of natural images from human brain activity},
  author={Naselaris, Thomas and Prenger, Ryan J and Kay, Kendrick N and Oliver, Michael and Gallant, Jack L},
  journal={Neuron},
  volume={63},
  pages={902--915},
  year={2009}
}

@InProceedings{gu2022decoding,
  title={Decoding natural image stimuli from fMRI data with a surface-based convolutional network},
  author={Gu, Zijin and Jamison, Keith and Kuceyeski, Amy and Sabuncu, Mert},
  booktitle ={MIDL},
  year ={2024}
}

@inproceedings{ozcelik2022reconstruction,
  title={Reconstruction of Perceived Images from fMRI Patterns and Semantic Brain Exploration using Instance-Conditioned GANs},
  author={Furkan, Ozcelik and Bhavin, Choksi and Milad, Mozafari and Leila, Reddy and Rufin, VanRullen},
  booktitle={IJCNN},
  year={2022},
}

@article{seeliger2018generative,
  title={Generative adversarial networks for reconstructing natural images from brain activity},
  author={Seeliger, Katja and G{\"u}c{\"u}l{\"u}, Umut and Ambrogioni, Luca and G{\"u}c{\"u}l{\"u}t{\"u}rk, Yagmur and van Gerven, Marcel AJ},
  journal={Neuroimage},
  volume={181},
  pages={775--785},
  year={2018}
}

@inproceedings{fang2020reconstructing,
  title={Reconstructing perceptive images from brain activity by shape-semantic GAN},
  author={Fang, Tao and Qi, Yu and Pan, Gang},
  booktitle={NeurIPS},
  year={2020}
}

@article{ren2021reconstructing,
  title={Reconstructing seen image from brain activity by visually-guided cognitive representation and adversarial learning},
  author={Ren, Ziqi and Li, Jie and Xue, Xuetong and Li, Xin and Yang, Fan and Jiao, Zhicheng and Gao, Xinbo},
  journal={Neuroimage},
  volume={228},
  year={2021}
}

@article{shen2019deep,
  title={Deep image reconstruction from human brain activity},
  author={Shen, Guohua and Horikawa, Tomoyasu and Majima, Kei and Kamitani, Yukiyasu},
  journal={PLoS Computational Biology},
  volume={15},
  year={2019}
}

@inproceedings{chen2023seeing,
  title={Seeing beyond the brain: Conditional diffusion model with sparse masked modeling for vision decoding},
  author={Chen, Zijiao and Qing, Jiaxin and Xiang, Tiange and Yue, Wan Lin and Zhou, Juan Helen},
  booktitle={CVPR},
  year={2023}
}

@article{ozcelik2023brain,
  title={Brain-diffuser:Natural scene reconstruction from fMRI signals using generative latent diffusion},
  author={Ozcelik, Furkan and VanRullen, Rufin},
  journal={Scientific Reports},
  volume={13},
  year={2023}
}

@inproceedings{huo2024neuropictor,
  title={Neuropictor: Refining fMRI-to-image reconstruction via multi-individual pretraining and multi-level modulation},
  author={Huo, Jingyang and Wang, Yikai and Qian, Xuelin and Wang, Yun and Li, Chong and Feng, Jianfeng and Fu, Yanwei},
  booktitle={ECCV},
  year={2024}
}

@inproceedings{quan2024psychometry,
  title={Psychometry: An Omnifit Model for Image Reconstruction from Human Brain Activity},
  author={Quan, Ruijie and Wang, Wenguan and Tian, Zhibo and Ma, Fan and Yang, Yi},
  booktitle={CVPR},
  year={2024}
}

@article{feifei2007perceive,
  title={What do we perceive in a glance of a real-world scene?},
  author={Fei-Fei, Li and Iyer, Asha and Koch, Christof and Perona, Pietro},
  journal={Journal of Vision},
  volume={7},
  number={1},
  pages={10--10},
  year={2007}
}

@article{brady2008visual,
  title={Visual long-term memory has a massive storage capacity for object details},
  author={Brady, Timothy F and Konkle, Talia and Alvarez, George A and Oliva, Aude},
  journal={Proceedings of the National Academy of Sciences of the United States of America},
  volume={105},
  pages={14325--14329},
  year={2008}
}

@article{hollingworth2001see,
  title={To see and remember: Visually specific information is retained in memory from previously attended objects in natural scenes},
  author={Hollingworth, Andrew and Williams, Christopher C and Henderson, John M},
  journal={Psychonomic Bulletin \& Review},
  volume={8},
  pages={761--768},
  year={2001}
}

@inproceedings{kim2020mixco,
  title={Mixco: Mix-up contrastive learning for visual representation},
  author={Kim, Sungnyun and Lee, Gihun and Bae, Sangmin and Yun, Se-Young},
  booktitle={NeurIPS Workshops},
  year={2020}
}

@inproceedings{rombach2022high,
  title={High-resolution image synthesis with latent diffusion models},
  author={Rombach, Robin and Blattmann, Andreas and Lorenz, Dominik and Esser, Patrick and Ommer, Bj{\"o}rn},
  booktitle={CVPR},
  year={2022}
}

@article{du2025human,
  title={Human-like object concept representations emerge naturally in multimodal large language models},
  author={Du, Changde and Fu, Kaicheng and Wen, Bincheng and Sun, Yi and Peng, Jie and Wei, Wei and Gao, Ying and Wang, Shengpei and Zhang, Chuncheng and Li, Jinpeng and others},
  journal={Nature Machine Intelligence},
  pages={1--16},
  year={2025},
}

@article{du2023decoding,
  title={Decoding visual neural representations by multimodal learning of brain-visual-linguistic features},
  author={Du, Changde and Fu, Kaicheng and Li, Jinpeng and He, Huiguang},
  journal={IEEE Transactions on Pattern Analysis and Machine Intelligence},
  volume={45},
  number={9},
  pages={10760--10777},
  year={2023},
}

@article{kamitani2005decoding,
  title={Decoding the visual and subjective contents of the human brain},
  author={Kamitani, Yukiyasu and Tong, Frank},
  journal={Nature neuroscience},
  volume={8},
  number={5},
  pages={679--685},
  year={2005},
}

@article{Orsborn2014closed,
author = {Orsborn, Amy and Moorman, Helene and Overduin, Simon and Shanechi, Maryam and Dimitrov, Dragan and Carmena, Jose},
year = {2014},
pages = {1380-93},
title = {Closed-Loop Decoder Adaptation Shapes Neural Plasticity for Skillful Neuroprosthetic Control},
volume = {82},
journal = {Neuron}
}

@article{chen2013ultrasensitive,
author = {Chen, Tsai-Wen and Wardill, Trevor and Sun, Yi and Pulver, Stefan and Renninger, Sabine and Baohan, Amy and Schreiter, Eric and Kerr, Rex and Orger, Michael and Jayaraman, Vivek and Looger, Loren and Svoboda, Karel and Kim, Douglas},
year = {2013},
pages = {295-300},
title = {Ultrasensitive fluorescent proteins for imaging neuronal activity},
volume = {499},
journal = {Nature}
}

@book{luck2011oxford,
  title={The Oxford handbook of event-related potential components},
  author={Luck, Steven J and Kappenman, Emily S},
  year={2011},
  publisher={Oxford university press}
}

@article{Roach2008EventrelatedET,
  title={Event-related EEG time-frequency analysis: an overview of measures and an analysis of early gamma band phase locking in schizophrenia.},
  author={Brian J. Roach and Daniel H. Mathalon},
  journal={Schizophrenia bulletin},
  year={2008},
  volume={34},
  pages={907--926}
}

@article{kam2005decoding,
author = {Kamitani, Yukiyasu and Tong, Frank},
year = {2005},
pages = {679-85},
title = {Decoding the visual and subjective contents of the human brain. Nature Neurosci.},
volume = {8},
journal = {Nature neuroscience}
}

@article{Horikawa2015GenericDO,
  title={Generic decoding of seen and imagined objects using hierarchical visual features},
  author={Tomoyasu Horikawa and Yukiyasu Kamitani},
  journal={Nature Communications},
  year={2015},
    pages={15037},
  volume={8}
}

@InProceedings{sen2020emerging,
author="Sen, Pratap Chandra
and Hajra, Mahimarnab
and Ghosh, Mitadru",
title="Supervised Classification Algorithms in Machine Learning: A Survey and Review",
booktitle={ETMG},
year="2020"
}

@INPROCEEDINGS{kumar2019bci,
  author={Kumar, Satyam and Yger, Florian and Lotte, Fabien},
  booktitle={BCI}, 
  title={Towards Adaptive Classification using Riemannian Geometry approaches in Brain-Computer Interfaces}, 
  year={2019}
}

@article{lotte2015electroencephalography,
  title={Electroencephalography (EEG)-based brain-computer interfaces},
  author={Lotte, Fabien and Bougrain, Laurent and Clerc, Maureen},
  journal={Wiley encyclopedia of electrical and electronics engineering},
  pages={44},
  year={2015},
}

@inproceedings{Wei2022ChainOT,
  title={Chain of Thought Prompting Elicits Reasoning in Large Language Models},
  author={Jason Wei and Xuezhi Wang and Dale Schuurmans and Maarten Bosma and Ed H. Chi and F. Xia and Quoc Le and Denny Zhou},
  booktitle={NeurIPS},
  year={2022},
}

@inproceedings{Kojima2022LargeLM,
  title={Large Language Models are Zero-Shot Reasoners},
  author={Takeshi Kojima and Shixiang Shane Gu and Machel Reid and Yutaka Matsuo and Yusuke Iwasawa},
  booktitle={NeurIPS},
  year={2022}
}

@article{yoshida2020natural,
  title={Natural images are reliably represented by sparse and variable populations of neurons in visual cortex},
  author={Yoshida, Takashi and Ohki, Kenichi},
  journal={Nature communications},
  volume={11},
  number={1},
  pages={872},
  year={2020},
}

@article{ozcelik2023natural,
  title={Natural scene reconstruction from fMRI signals using generative latent diffusion},
  author={Ozcelik, Furkan and VanRullen, Rufin},
  journal={Scientific Reports},
  volume={13},
  number={1},
  pages={15666},
  year={2023},

}

@article{miyawaki2008visual,
  title={Visual image reconstruction from human brain activity using a combination of multiscale local image decoders},
  author={Miyawaki, Yoichi and Uchida, Hajime and Yamashita, Okito and Sato, Masa-aki and Morito, Yusuke and Tanabe, Hiroki C and Sadato, Norihiro and Kamitani, Yukiyasu},
  journal={Neuron},
  volume={60},
  number={5},
  pages={915--929},
  year={2008},
}

@article{cowen2014neural,
  title={Neural portraits of perception: reconstructing face images from evoked brain activity},
  author={Cowen, Alan S and Chun, Marvin M and Kuhl, Brice A},
  journal={Neuroimage},
  volume={94},
  pages={12--22},
  year={2014},
}

@article{li2025neuraldiffuser,
  title={NeuralDiffuser: Neuroscience-Inspired Diffusion Guidance for fMRI Visual Reconstruction},
  author={Li, Haoyu and Wu, Hao and Chen, Badong},
  journal={IEEE Transactions on Image Processing},
  year={2025},
}

@inproceedings{daimindaligner,
  title={MindAligner: Explicit Brain Functional Alignment for Cross-Subject Visual Decoding from Limited fMRI Data},
  author={Dai, Yuqin and Yao, Zhouheng and Song, Chunfeng and Zheng, Qihao and Mai, Weijian and Peng, Kunyu and Lu, Shuai and Ouyang, Wanli and Yang, Jian and Wu, Jiamin},
  booktitle={ICML},
year={2025}
}

@inproceedings{gong2025mindtuner,
  title={Mindtuner: Cross-subject visual decoding with visual fingerprint and semantic correction},
  author={Gong, Zixuan and Zhang, Qi and Bao, Guangyin and Zhu, Lei and Xu, Rongtao and Liu, Ke and Hu, Liang and Miao, Duoqian},
  booktitle={AAAI},
  year={2025}
}

@inproceedings{lu2023minddiffuser,
  title={Minddiffuser: Controlled image reconstruction from human brain activity with semantic and structural diffusion},
  author={Lu, Yizhuo and Du, Changde and Zhou, Qiongyi and Wang, Dianpeng and He, Huiguang},
  booktitle={ACMMM},
  year={2023}
}

@inproceedings{xu2023versatile,
  title={Versatile diffusion: Text, images and variations all in one diffusion model},
  author={Xu, Xingqian and Wang, Zhangyang and Zhang, Gong and Wang, Kai and Shi, Humphrey},
  booktitle={ICCV},
  year={2023}
}

@inproceedings{kavasidis2017brain2image,
  title={Brain2image: Converting brain signals into images},
  author={Kavasidis, Isaak and Palazzo, Simone and Spampinato, Concetto and Giordano, Daniela and Shah, Mubarak},
  booktitle={ACM MM},
  year={2017}
}

@article{jin2024pgcn,
  title={PGCN: Pyramidal graph convolutional network for EEG emotion recognition},
  author={Jin, Ming and Du, Changde and He, Huiguang and Cai, Ting and Li, Jinpeng},
  journal={IEEE Transactions on Multimedia},
  volume={26},
  pages={9070--9082},
  year={2024},
}

@inproceedings{davis2022brain,
  title={Brain-supervised image editing},
  author={Davis, Keith M and De La Torre-Ortiz, Carlos and Ruotsalo, Tuukka},
  booktitle={CVPR},
  year={2022}
}

\vfill

\end{document}